# Deep Learning for Model Calibration in Simulation of Itaconic Acid Production

Daria Fokina[1], Marco Baldan[1], Constantin Romankiewicz[1], Wolfgang Laudensack[2], Roland Ulber[2], Michael Bortz[1*]

[1]Fraunhofer ITWM, Kaiserslautern, Germany

[2]RPTU Kaiserslautern, Kaiserslautern, Germany

[*]Corresponding author. Email: michael.bortz@itwm.fraunhofer.de

## Abstract

In this study, deep learning is used to estimate kinetic parameters for modeling itaconic acid production based on real batch experiments conducted at different agitation speeds and reactor scales. Two deep learning strategies, namely direct deep learning (DDL) and generative conditional flow matching (CFM) are compared and benchmarked against nonlinear regression as a reference method. Compared with DDL, CFM consistently yields more accurate results. The concentration profiles predicted by CFM closely match those obtained from nonlinear regression, whereas DDL results in larger deviations. Similar behavior is observed in the scale-up experiments, where the CFM model again generalizes better and is more robust than the direct approach. These findings demonstrate that CFM can reliably predict system behavior across different operating conditions and scales, offering a flexible and data-efficient framework for parameter estimation in dynamic bioprocess models.

**Key words:** Itaconic acid modeling, delay differential equations, deep learning, conditional flow matching.

## 1 Introduction

Itaconic acid is a promising bio-based platform chemical produced by microbial fermentation. In recent years, increased research efforts have focused on understanding and optimizing its production by the smut fungus *Ustilago maydis* (Becker et al. 2020). This yeast-like, unicellular growing fungus synthesizes itaconic acid from carbon sources by redirecting flux from the tricarboxylic acid cycle towards acid secretion (Geiser et al. 2016).

In *U. maydis*, exponential growth continues until nitrogen becomes limiting, at which point itaconic acid production is initiated (Klement et al. 2012). The acid is secreted into the extracellular medium and accumulates during the production phase. While a few continuous process setups have been reported (Carstensen et al. 2013), most works focus on batch and fed-batch strategies, including pulsed fed-batch approaches (Helm et al. 2024). The highest reported titers reached 220 g·L$^{-1}$ under optimized fed-batch conditions (Hosseinpour Tehrani et al. 2019), demonstrating the industrial relevance of this production system.

Several studies have investigated metabolic fluxes and genome-scale models of *U. maydis* for itaconic acid production (Voll et al. 2012; Liebal et al. 2022; Ziegler et al. 2024). These works primarily address intracellular metabolism and do not explicitly capture the macroscopic dynamics of biomass, substrate, and product concentrations in batch cultures. Quantitative descriptions of these process-level interactions remain limited, highlighting the need for modeling approaches suitable for scale-up and process engineering applications.

Model-based simulation is a promising strategy for process optimization. Therefore, accurate mathematical models are needed that reliably describe bioprocesses like fermentation. Unstructured models, which focus on macroscopic variables such as biomass, substrate, and product concentration, are widely used in chemical and bioprocess engineering due to their relative simplicity and lower parameter requirements compared to their structured counterparts (Villadsen et al. 2011). However, a significant limitation of standard Ordinary Differential Equation (ODE)-based unstructured models is their inherent assumption of instantaneous cellular response, assuming all system responses are immediate. This means that the rate of growth or product formation at time $t$ depends only on the environmental conditions present at $t$.

However, any fundamental microbial and cellular processes such as cell growth, nutrient storage, and product formation, are inherently non-instantaneous processes that involve a sequence of biochemical and transport steps. This lag between environmental change, such as substrate uptake, and the corresponding cellular response, such as biomass growth or product secretion, introduces a time delay or memory into the system dynamics. To address such time delay dynamics, Delay Differential Equations (DDEs), also known as functional differential equations, have been introduced (Rihan et al. 2018) and are used in this work. These equations explicitly incorporate the system's history by allowing the derivative of a state variable to depend on its past values. Furthermore, they can be seen as a shortcut model to capture dependencies that, due to lack of data, are not modeled explicitly (Glass et al. 2021). A state-dependent DDE for a variable $X$ can be generally expressed as:

$$\frac{dX(t)}{dt} = f\big(X(t), X(t - \tau_1), \ldots, X(t - \tau_m)\big) \, , \tag{1}$$

where $\tau_1, \ldots, \tau_m$ are the time delay terms.

The function $f$ in Equation (1) is a parametric function whose parameters depend on the process conditions. The conventional approach to estimating these parameters is nonlinear regression (Bates and Watts 1988; Biegler 2010). If convergence to the global optimum is reached, it yields highly accurate benchmark results. However, this method can be sensitive to initialization, convergence of the optimization solver and prone to local minima. It may struggle when the relationships between parameters and process variables are highly non-linear, or when there are several distinct parameter sets that can equally well explain the data. Therefore, in this contribution, deep learning methods that circumvent dependencies on starting values and convergence issues are considered and benchmarked against nonlinear regression results.

Two deep learning methods are considered: Direct deep learning (DDL) (Rensonnet et al. 2021; Raponi and Marchisio 2024) and conditional flow matching (CFM) (Sherki et al. 2025). For both methods, a set of training data is created by a high number of model evaluations for different model parameters and initial conditions. In DDL, a deep neural network is trained with the model predictions and initial conditions as inputs and the model parameters as outputs.

CFM works differently: Its use for parameter estimation has been suggested in (Sherki et al. 2025). By leveraging ideas from generative artificial intelligence, the conditional distribution of parameters is obtained, rather than a single point estimate. In this sense, it is comparable to Bayesian parameter estimation (BPE) (Wulkow et al. 2021). However, whereas the sampling strategy underlying BPE depends on the measured experimental data, the conditional flow in CFM is universal for the model; the measured data enter only at the very end as its input.

In the present work, based on real experimental data of itaconic acid production, it is demonstrated that, compared to DDL, CFM can better capture complex parameter-process relationships, provide uncertainty estimates, and potentially improve numerical robustness when combined with traditional nonlinear regression.

The remainder of this paper is organized as follows: In section two, details on the model and parameter estimation strategies are given, including nonlinear regression, DDL and CFM. Section three deals with the experimental setup. In section four, results are presented for the data and models at hand. The paper ends with a summary and outlook.

## 2 Modeling and Model Calibration

### 2.1 Model Calibration via Nonlinear Regression

To describe the dynamic behavior of itaconic acid production in a stirred-tank bioreactor, we developed a delay differential equation (DDE) unstructured segregated model incorporating biomass growth, substrate (i.e. glucose) consumption, and product formation kinetics. The formulation extends classical Monod-type (Monod 1949) models by including a single, uniform discrete time delay, $\tau_S$, representing the intracellular metabolic lag between substrate uptake and itaconic acid biosynthesis. The model tracks three primary state variables: $X$, viable biomass concentration ($g\ L^{-1}$), $S$, limiting carbon substrate concentration ($g\ L^{-1}$), glucose in this case, and $P$, itaconic acid concentration ($g\ L^{-1}$). The reactor is operated in batch conditions; therefore, dilution terms are omitted.

Biomass growth is described here by Monod kinetics:

$$\frac{dX(t)}{dt} = -k_d X + \mu_m \frac{S(t)}{S(t) + K_S}, \tag{2}$$

where $k_d$ is the dead rate, $\mu_m$ is the maximum specific growth rate, $K_S$ is the Monod half-saturation constant. Substrate is consumed for biomass synthesis and maintenance. The substrate balance is given by:

$$\frac{dS(t)}{dt} = -Y_{XS}^{-1} \mu_m \frac{S(t)}{S(t) + K_S} X, \tag{3}$$

where $Y_{XS}$ is the biomass yield on substrate. Finally, itaconic acid formation is modeled as a delayed-substrate-associated process:

$$\frac{dP(t)}{dt} = k_p \frac{S(t - \tau_S)}{S(t - \tau_S) + K_{PS}} \frac{X(t)}{X(t) + K_X}, \tag{4}$$

where $k_p$ is the product formation coefficient, $K_{PS}$ and $K_X$ are Monod half-saturation constants. The initial history of the functions $X(t), S(t), P(t)$ for $t \in [-\tau_S, 0]$ is taken to be constant and equal to the initial values.

As usual in nonlinear regression, the model parameters are adjusted to each experiment individually via minimization of weighted sum of squared errors:

$$\min_{k_d,\mu_m,K_S,Y_{XS},k_p,t_S,K_{PS},K_X} SSE$$

$$\text{with } SSE = w_X \sum_t^a \left(X_{predicted}^i - X_{data}^i\right)^2 + w_S \sum_t^a \left(S_{predicted}^i - S_{data}^i\right)^2$$

$$+ w_P \sum_t^a \left(P_{predicted}^i - P_{data}^i\right)^2, \quad (5)$$

where $w_X, w_S$ and $w_P$ are weights in order to normalize $X, S$ and $P$.

For model implementation and subsequent optimization, Julia (Rackauckas and Nie 2017) and IPOpt (Biegler and Zavala 2009) were used, including calculation of exact gradients. Solving (5) for one experiment requires about 1000 forward evaluations of the DDE solver. The optimization bounds for optimized parameters are defined in Table 1.

*Table 1. Model parameter bounds, used for conventional regression and data generation for deep learning methods.*

| Parameter | $k_d$ | $\mu_m$ | $K_S$ | $Y_{XS}^{-1}$ | $k_p$ | $\tau_S$ | $K_{PS}$ | $K_X$ |
|---|---|---|---|---|---|---|---|---|
| Unit | $[h^{-1}]$ | $[h^{-1}]$ | $[g/L]$ | $[L/g]$ | $[g/(L \cdot h)]$ | $[h]$ | $[g/L]$ | $[g/L]$ |
| Lower bound | $10^{-3}$ | 0.1 | 10 | 0.1 | $5 \times 10^{-3}$ | 10 | 50 | 30 |
| Upper bound | $10^{-2}$ | 1 | 100 | 10 | 0.3 | 50 | 70 | 100 |

We used multi-start routines in order to ensure robust convergence to the global optimum. This yields highly accurate results that will be used as a benchmark for the deep learning approaches.

## 2.2  Model Calibration Using Deep Learning Methods

To avoid globality and convergence issues encountered in nonlinear regression, in this work deep learning is explored as an alternative method for parameter estimation. Deep learning models are widely known for their ability to approximate complex nonlinear functions (Hornik et al. 1989). In our case, the goal is to find a function $f$ with:

$$f: \eta \to \xi,$$

which finds model parameters $\xi = (k_d, \mu_m, K_S, Y_{XS}^{-1}, k_p, \tau_s, K_{PS}, K_X)$ for measured time and concentration profiles $\eta = (t, X, S, P)$, considered as inputs.

To approximate $f$ two methods are considered here: a straightforward DDL approach, which predicts the parameters directly from the measurements, and a CFM approach, which first estimates the velocity field between two distributions and then obtains the parameters via solving the underlying stochastic ODE.

### 2.2.1  DDL Approach

The DDL approach implies learning the model parameters directly from the measurements. This approach uses a parametric model to map the measured trajectories directly to the biokinetics parameters. The model is represented as a neural network $NN(\eta; \theta)$. To find the neural network parameters $\theta$, the mean squared error loss function is optimized:

$$\min_{\theta} \frac{1}{N_{data}} \sum_{i=1}^{N_{data}} \|NN(\eta_i; \theta) - \xi_i\|^2 \quad (6)$$

where $N_{data}$ is the number of points in generated training data. After training, to obtain the biokinetics parameters given experimental measurements, only a computationally cheap direct evaluation of a neural network is required.

### 2.2.2 CFM Approach

CFM is a generative approach modeling probability paths for unknown data distributions. The training target is a velocity field that generates a conditional probability path (Lipman et al. 2023) . Originally applied in image generation (Peebles and Xie 2023), it demonstrated capability to solve Bayesian inverse problems (Sherki et al. 2025).

The goal of these problems is to estimate a posterior distribution $p(\xi|\eta)$, where $\xi$ are unobservable model parameters, $\eta$ are experimental observations. The CFM objective is to learn the velocity field that transports a simple prior distribution $p_0$ to the conditional posterior distribution $p$. The transport dynamics is given by a (stochastic) ordinary differential equation:

$$\frac{d\xi_\tau}{d\tau} = v(\xi_\tau, \tau),$$

with initial condition $\xi_0 \sim p_0$, $\tau$ is the artificial time variable, $\xi_1 = \xi$ are the desired model parameters.

If a linear path between prior distribution $p_o$ and target distribution $p$ is selected, meaning $\xi_\tau = (1-\tau)\xi_0 + \tau\xi_1$, then the velocity field $v$ is constant along the path and equal to

$$v = \xi_1 - \xi_0.$$

To approximate the velocity field by a neural network, the optimization problem is solved:

$$\min_{\theta} \mathbb{E}_{\tau,\xi_0,(\xi,\eta,e)\sim\rho}[\|NN(\xi_\tau, \tau, \eta; \theta) - (\xi - \xi_0)\|].$$

Once trained, the CFM model, together with experimental observations, defines a velocity field that can be integrated using standard ODE solvers to infer the parameters of the underlying mathematical process model (Lipman et al. 2023).

In the field of deep learning, choosing the right neural network architecture is essential as it can greatly impact the performance of the model. In the last years, transformer networks (Vaswani et al. 2017) have grown to be one of the predominant deep neural network architectures. While they were first and foremost applied to language modeling tasks, they have since proven valuable in other domains. CFM has been successfully used with a transformer network architecture to solve Bayesian inverse problems (Sherki et al. 2025). This architecture enables the handling of arbitrary input lengths, allowing experiments with different numbers of time-point observations.

This work focuses on the use of a transformer model for obtaining model parameters within the CFM approach. The model follows the architecture of (Sherki et al. 2025) with a 32-dimensional projection space and ten 8-head attention layers, resulting in approximately $1.37 \times 10^5$ parameters in total. Further studies investigating the use of multilayer Perceptron (MLP) can be found in Additional file 1 (Section 1).

## 2.3 Data Generation

Both deep learning methods considered here require the same training data (~1000 samples). For itaconic acid modeling, where one experiment takes several days, such amounts of data are not available. Therefore, a synthetic dataset was generated using randomly selected parameter values and initial substrate concentration. In total 10000 samples were generated via Latin hypercube sampling, with the same ranges as used in conventional regression, see Table 1. The initial substrate concentrations were chosen from the range [50, 60] g/L. Concentration values of biomass, substrate and product were obtained by solving Equations (2), (3) and (4) for randomly selected time points. The generated dataset was split into two subsets: a training set of up to 5000 samples and a test set of 2000 samples. As their names suggest, the training set was used to fit the neural network parameters and the test set was used for evaluation.

The model quality is evaluated using the normalized root mean squared error (root mean squared error, divided by the scale):

$$NRMSE = \frac{\sqrt{1/N \sum_i^N (y_i - \hat{y}_i)^2}}{\max_i y_i - \min_i y_i} \qquad (7)$$

where $y_i$ are the data points, $\hat{y}_i$ are the model predictions, $N$ is the number of samples. The error is computed separately for each predicted model parameter (in which case $\hat{y}_i = \xi_i$) and for the corresponding concentration profiles (in which case $\hat{y}_i = \eta_i$).

Finally, these two approaches were applied to the experimental data.

## 3 Experimental Setup

The dataset used for model development comprised newly generated experimental data as well as previously published data obtained during screening of agitation speeds and aeration rates (Volkmar et al. 2025). While a detailed description of the experimental procedures is provided in the original publication, a concise summary of the most relevant materials and methods is given below.

The employed production strain was the genetically modified *Ustilago maydis* MB215 ΔCyp3 $P_{etef}$Ria1, constructed and kindly provided by the working group of Blank (RWTH Aachen, Germany) (Hosseinpour Tehrani et al. 2019). Batch fermentations were performed in the 0.5 L Biostat Q+ reactor system (Sartorius, Göttingen, Germany) with a working volume of 0.4 L of modified Tabuchi medium. The medium contains 19.5 g·L$^{-1}$ MES buffer, 50 g·L$^{-1}$ glucose, 0.8 g·L$^{-1}$ ammonia chloride, 0.5 g·L$^{-1}$ monopotassium phosphate, 0.2 g·L$^{-1}$ magnesium sulfate heptahydrate, 0.1 g·L$^{-1}$ iron sulfate heptahydrate, vitamins and trace elements at pH 6.5 (Geiser et al. 2014). Reactors were inoculated to an initial $OD_{600}$ of 0.5 from overnight cultures and aerated at 1 or 2.5 volumes of air per volume of medium per minute (vvm).

For scale-up experiments, batch fermentations were conducted in a 42 L Biostat C+ reactor system (Sartorius, Göttingen, Germany) with a working volume of 30 L and an aeration rate of 1 vvm under otherwise similar conditions as described for the 0.5 L reactor.

Biomass concentration was monitored by optical density at 600 nm ($OD_{600}$) with CO8000 Cell Density Meters (Biochrom Ltd., Cambridge, UK) and converted to biomass concentration using device-specific correlations ($OD_{600}$ 1 = 0.2578 $g_{BM}$·L$^{-1}$ or 0.2040 $g_{BM}$·L$^{-1}$). Substrate and product concentrations were determined by HPLC using a refractive index detector (RI 101 Shodex, Showa Denke KK, Tokyo, Japan), with separation on a Bio-Rad Aminex HPX-87H column (300 × 7.8 mm, Hercules, California, USA) at 80°C and 2.5 mM $H_2SO_4$ as the mobile phase at a flow rate of 0.6 mL·min$^{-1}$.

# 4 Results and Discussion

## 4.1 Deep Learning Models Fitting on Synthetic Data

The direct and CFM deep-learning approaches were initially trained on artificially generated data. Their training results are examined in this section.

### 4.1.1 DDL Approach

First, the DDL approach is considered, which includes learning the direct mapping between the values $(t, X, S, P)$ and the model parameters $(k_d, \mu_m, K_S, Y_{XS}^{-1}, k_p, \tau_s, K_{PS}, K_X)$.

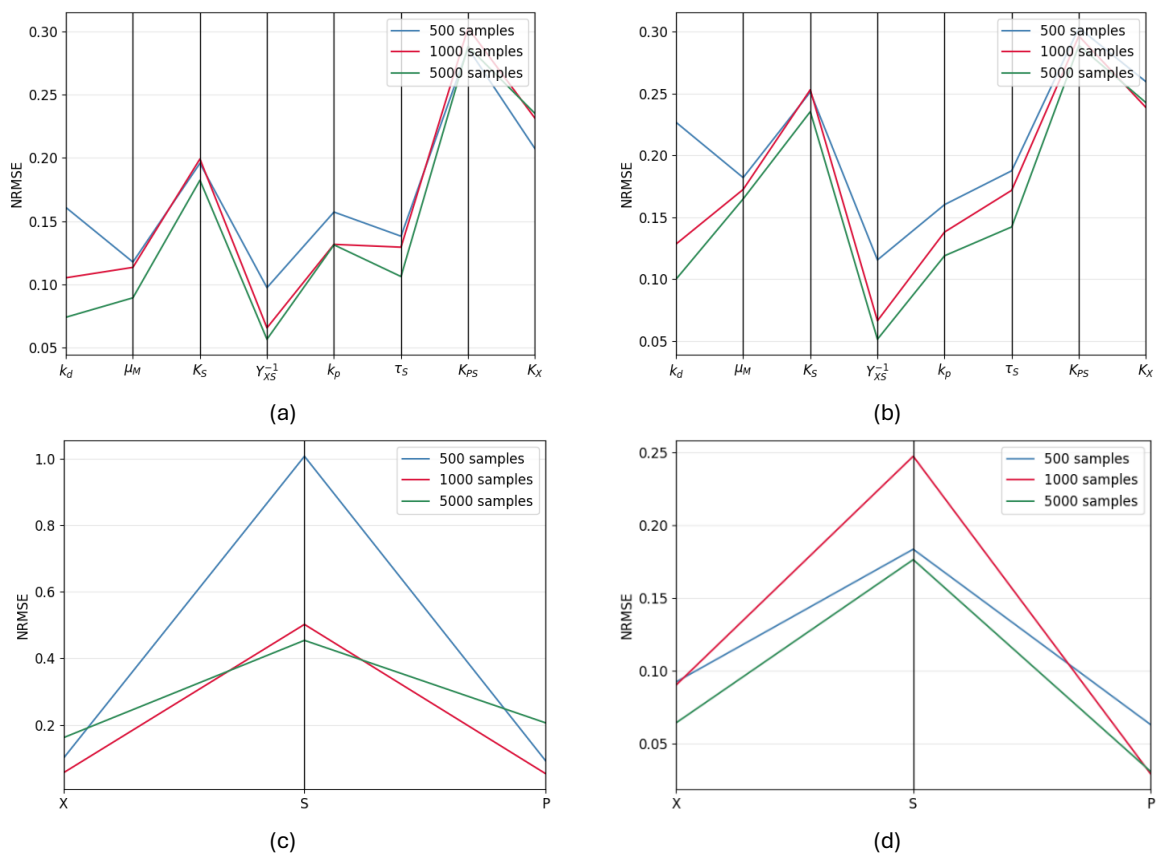

*Figure 1. Parallel coordinates plots for errors for direct deep learning approach.*
*Parallel coordinates plots show errors of the direct deep learning approach in model parameter predictions on train (a) and test (b) sets and corresponding solutions on the train (c) and test (d) sets. $k_d, \mu_m, K_S, Y_{XS}^{-1}, k_p, \tau_s, K_{PS}, K_X$ are model parameters, X (biomass concentrations), S (glucose concentration), P (itaconic acid concentration) are the DDE solutions. The predictions were obtained using model trained on 500 (blue line), 1000 (red line) and 5000 (green line) samples.*

To start with, the influence of the number of samples on the overall performance of the neural network is studied. The normalized root mean squared errors for 500, 1000 and 5000 training samples are presented in Figure 1. Figures (a) and (b) show the errors in the predicted model parameters. Once the model parameters have been determined, the underlying system of delayed differential equations (Equations (2)-(4) ) can be solved, and the resulting solutions can be compared with the given data. Figures (c) and (d) show the NRMSE values computed for the DDE solutions.

When 500 samples were used, a few samples predicted negative values for the biomass yield on the substrate parameter, $Y_{XS}^{-1}$. Figure 1 shows the error values, when excluding predictions with negative values of this parameter.

### 4.1.2 CFM Approach

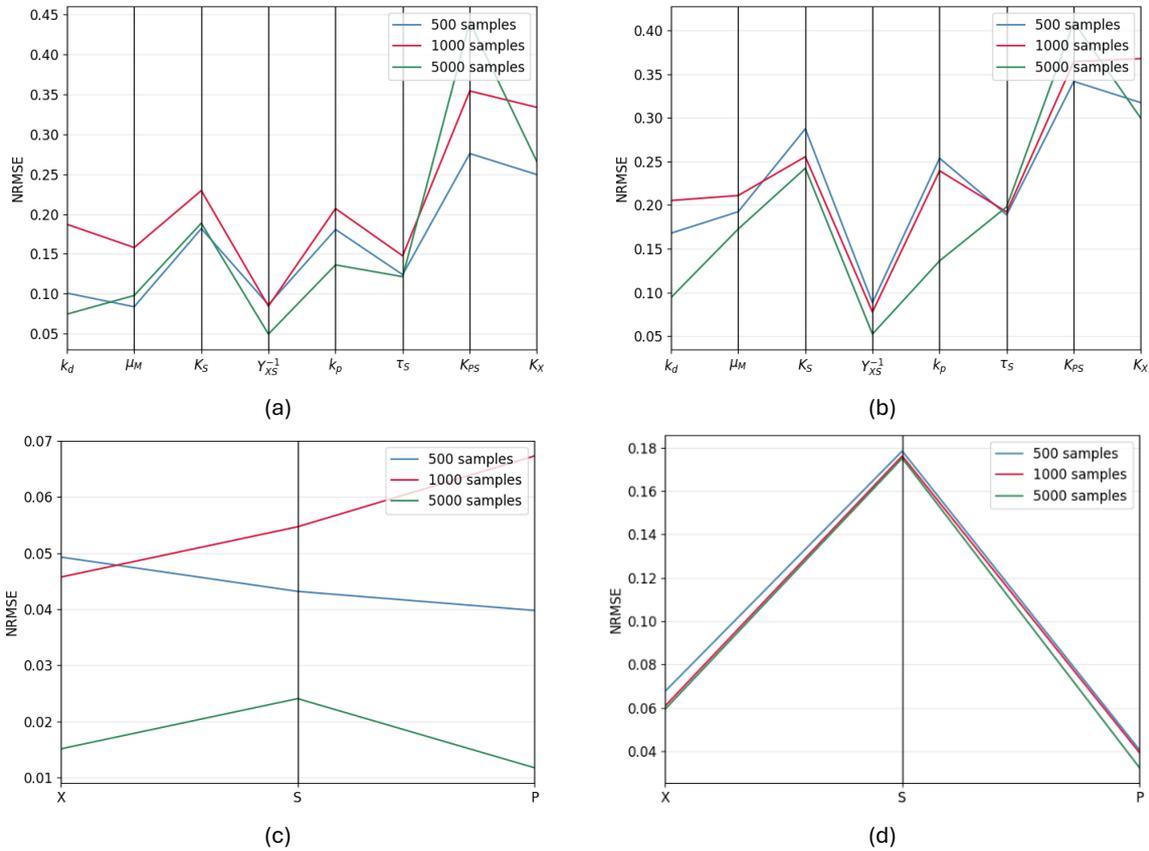

*Figure 2. Parallel coordinates error plots for CFM approach.*
*Parallel coordinates plots show errors of the direct deep learning approach in model parameter predictions on the train (a) and the test (b) sets and corresponding solutions on the train (a) and the test (b) sets for CFM approach. $k_d, \mu_m, K_S, Y_{XS}^{-1}, k_p, \tau_s, K_{PS}, K_X$ are model parameters, X (biomass concentrations), S (glucose concentration), P (itaconic acid concentration) are the DDE solutions. The predictions were obtained using model trained on 500 (blue line), 1000 (red line) and 5000 (green line) samples.*

A similar analysis was performed for the CFM algorithm. Figure 2 presents the errors on both the training and test datasets for different numbers of training samples (500, 1000, and 5000). The CFM approach delivers performance comparable to that of the direct model for all dataset sizes. Remarkably, the CFM model achieves highly accurate results with only 500 training samples, demonstrating its efficiency in learning data structures. As the size of the training dataset increases, the model's accuracy remains consistently high, improving only slightly. This indicates that the CFM model is robust and does not depend heavily on large amounts of training data.

## 4.2 Parameter estimation on experimental data

Experiments were conducted in a 0.5 L reactor at aeration rates of 2.5 vvm and 1 vvm, and in a 42 L reactor at an aeration rate of 1 vvm. Measurements were taken at multiple agitation speeds. The measured data included the concentrations of biomass, glucose, and itaconic acid at several time points. Each concentration profile corresponds to a single agitation speed. The results presented below show only a subset of the data.

### 4.2.1 Results for experiments in 0.5 L reactor at aeration rate 2.5 vvm

The solution for the parameters generated by the direct deep learning approach for experimental data at an agitation speed of 100 rpm is presented in Figure 3. Figure 4 shows the solution for CFM predictions. More examples are provided in Sections 2.2 and 2.3 of Additional file 1. The summary table for predicted biokinetics parameters is provided in Table 2.

*Table 2. Parameter values obtained for experimental data (100 and 600 rpm, 1 vvm, 0.5 L reactor).*

|  | Parameter | $k_d$ [$h^{-1}$] | $\mu_m$ [$h^{-1}$] | $K_S$ [$g/L$] | $Y_{XS}^{-1}$ [$L/g$] | $k_p$ [$g/(L \cdot h)$] | $\tau_s$ [$h$] | $K_{PS}$ [$g/L$] | $K_X$ [$g/L$] |
|---|---|---|---|---|---|---|---|---|---|
| 100 rpm | Regression | 0.0082 | 0.5273 | 100.0 | 1.361 | 0.1381 | 32.57 | 61.53 | 62.5 |
|  | DDL | 0.0092 | 0.5392 | 81.85 | 1.388 | 0.1665 | 24.74 | 59.23 | 57.57 |
|  | CFM | 0.008 | 0.4622 | 79.89 | 1.532 | 0.1787 | 26.65 | 61.74 | 72.83 |
| 600 rpm | Regression | 0.0065 | 0.3352 | 10.5 | 0.945 | 0.0941 | 37.57 | 61 | 50.56 |
|  | DDL | 0.0058 | 0.518 | 37.52 | 1.196 | 0.0822 | 38.66 | 61.88 | 64.83 |
|  | CFM | 0.0057 | 0.488 | 29.27 | 1.045 | 0.0779 | 45.52 | 61.8 | 75.9 |

*The parameters were obtained using conventional regression, direct deep learning approach and conditional flow matching.*

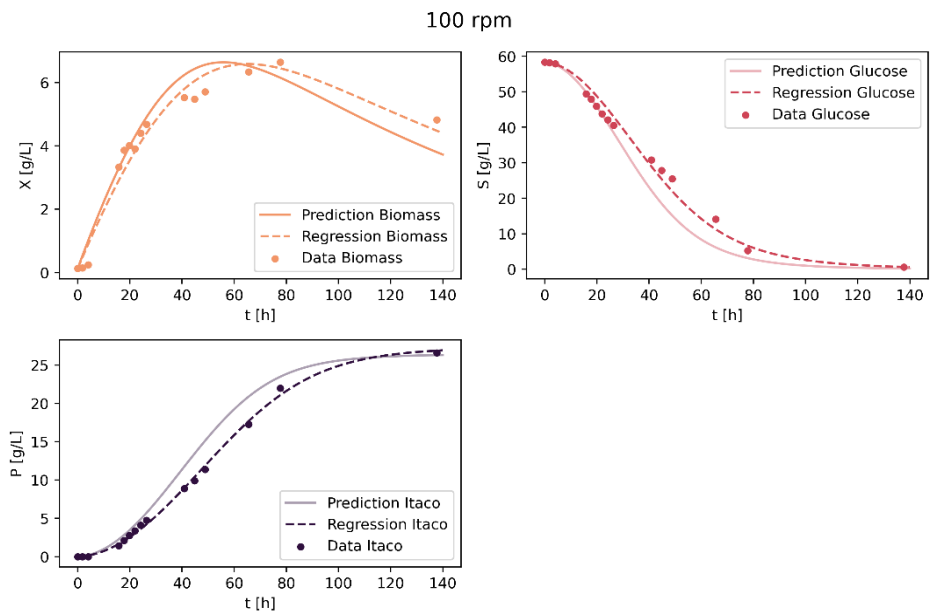

*Figure 3. Solutions for parameters predicted by direct deep learning approach vs experimental data.*
*Solutions correspond to the operating conditions with a mixer speed of 100 rpm, an aeration rate of 2.5 vvm in a 0.5 L reactor.*

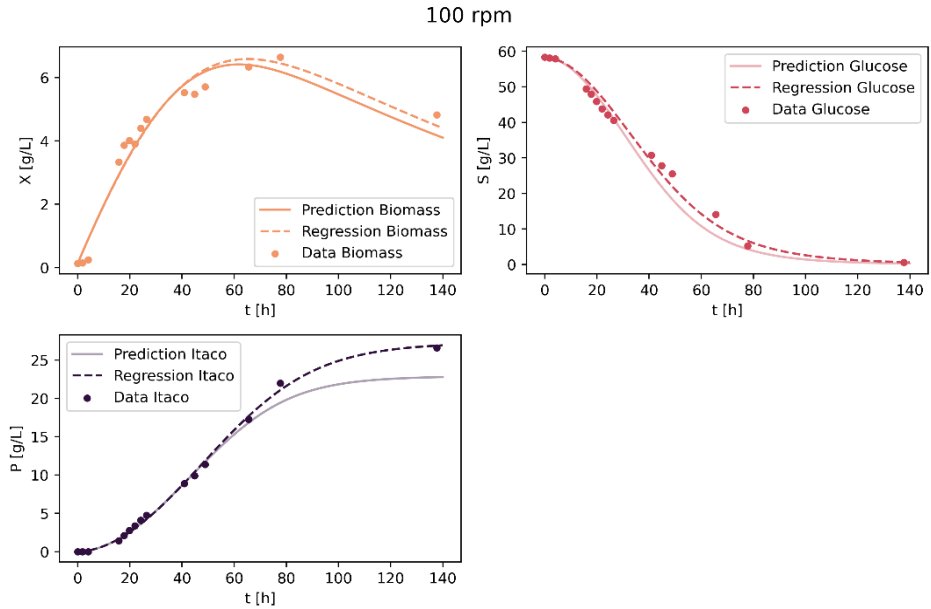

*Figure 4. Solutions for parameters predicted by CFM vs experimental data.*
*Solutions correspond to the operating conditions with a mixer speed of 100 rpm, an aeration rate of 2.5 vvm in a 0.5 L reactor.*

Table 3 shows the comparison of normalized mean squared error values for three considered approaches. As the results show, the conditional flow matching method produces results like those of conventional regression, but with notably lower error rates than the direct deep-learning model. This indicates that CFM provides a more accurate and robust estimation of biokinetic parameters, also when applied to experimental datasets. Consistently lower error values demonstrate that CFM can capture underlying system dynamics more reliably than the DDL approach.

A major advantage of the CFM method is that it provides a full distribution of plausible parameter values, rather than just single parameter estimates. This enables parameter uncertainty to be quantified directly from the model outputs. The corresponding uncertainty estimates for the experimental data are shown in Additional file 1 (Section 2.3).

*Table 3. Normalized mean squared error in predictions.*

| Experiment | Error | | | | | | | | |
| --- | --- | --- | --- | --- | --- | --- | --- | --- | --- |
| | Biomass | | | Glucose | | | Itaconic acid | | |
| | Regression | DDL | CFM | Regression | DDL | CFM | Regression | DDL | CFM |
| 100 rpm | 0.062 | 0.085 | 0.067 | 0.04 | 0.081 | 0.056 | 0.015 | 0.072 | 0.053 |
| 400 rpm | 0.048 | 0.133 | 0.064 | 0.023 | 0.096 | 0.033 | 0.03 | 0.068 | 0.079 |
| 600 rpm | 0.069 | 0.114 | 0.088 | 0.039 | 0.039 | 0.037 | 0.027 | 0.074 | 0.028 |

*Predictions were obtained using nonlinear regression, direct deep learning approach and CFM approach for experiments in small reactor at an aeration rate 2.5 vvm and mixer speeds of 100, 400 and 600 rpm in a 0.5 L reactor.*

*Table 4. Parameter values obtained for experimental data (300 and 600 rpm, 1 vvm, 0.5 L reactor).*

|  | Parameter | $k_d$ [$h^{-1}$] | $\mu_m$ [$h^{-1}$] | $K_S$ [$g/L$] | $Y_{XS}^{-1}$ [$L/g$] | $k_p$ [$g/(L \cdot h)$] | $\tau_s$ [$h$] | $K_{PS}$ [$g/L$] | $K_X$ [$g/L$] |
|---|---|---|---|---|---|---|---|---|---|
| 300 rpm | Regression | 0.0034 | 0.2889 | 100.0 | 2.607 | 0.0533 | 49.25 | 50.0 | 30.0 |
|  | DDL | 0.0054 | 0.3346 | 94.44 | 3.298 | 0.1389 | 41.99 | 59.45 | 46.61 |
|  | CFM | 0.0021 | 0.2524 | 92.83 | 3.261 | 0.0559 | 48.94 | 57.73 | 38.88 |
| 600 rpm | Regression | 0.0072 | 0.3285 | 10.0 | 0.915 | 0.1047 | 23.94 | 60.53 | 36.86 |
|  | DDL | 0.007 | 0.3702 | 15.47 | 1.779 | 0.1487 | 22.49 | 60.68 | 55.96 |
|  | CFM | 0.0064 | 0.4475 | 26.6 | 1.048 | 0.1194 | 26.36 | 58.43 | 75.11 |

*The parameters were obtained using conventional regression, direct deep learning approach and conditional flow matching.*

### 4.2.2 Results for experiments in 0.5 L reactor at aeration rate 1 vvm

Several experiments were conducted at aeration rate of 1 vvm. Table 4 shows the predicted parameter values for measurements taken at agitation speeds of 300 and 600 rpm. The lowest considered speed is 300 rpm. The results demonstrate that nonlinear regression, the direct deep learning model, and the CFM approach produce different coefficient estimates. The concentration profiles at 300 rpm (Figure 5 for DDL and Figure 6 for the CFM) show that CFM provides considerably better predictions and closely aligns with the nonlinear regression results.

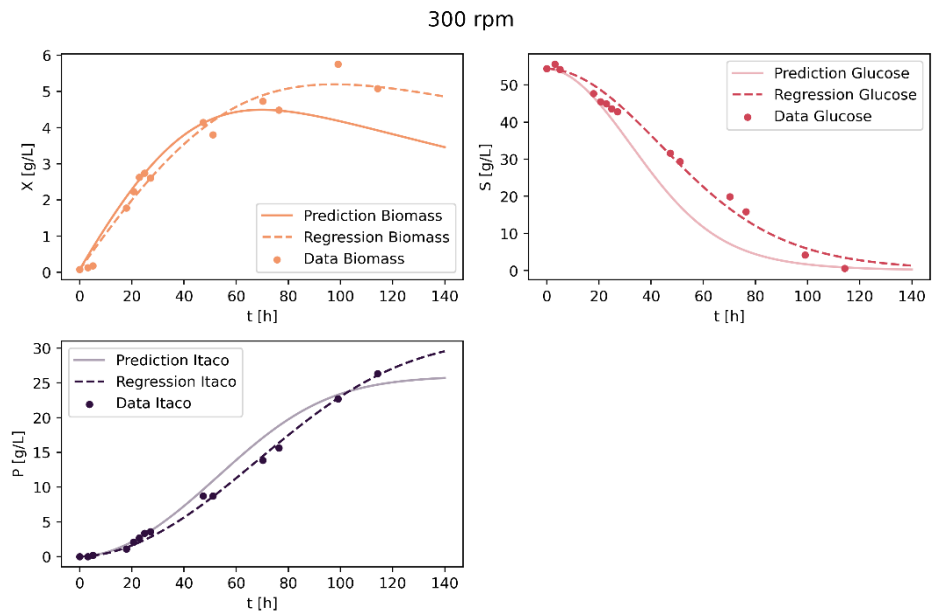

*Figure 5. Comparison of direct deep learning approach result (solid line) with conventional regression (dashed line). Solutions correspond to the operating conditions with a mixer speed of 300 rpm, an aeration rate of 1 vvm in a 0.5 L reactor.*

### 4.2.3 Results for experiments in 42 L reactor at aeration rate 1 vvm

The experiments were also conducted in a 42 L scale-up reactor with a working volume of 30 L at aeration rate of 1 vvm. An agitation speed of 114 rpm corresponds to a scale-up based on constant impeller tip speed from 300 rpm in the 0.5 L reactor, while 229 rpm represents a scale-up over impeller tip speed from 600 rpm in the 0.5 L reactor, as described in (Volkmar et al. 2025). Table 5 shows the parameter estimates obtained at agitation speeds of 114 and 229 rpm.

As in previous cases, the direct deep learning model and the CFM method produce different sets of coefficients.

The concentration profiles for 114 rpm are provided in Figure 7 for the direct model and Figure 8 for the CFM model. The figures demonstrate that the CFM approach accurately reproduces the system dynamics, showing strong agreement with the nonlinear regression results. In contrast, the DDL model shows more substantial deviations from the expected behavior. As with previous experiments, these results show that the CFM method outperforms the DDL model in terms of predictive accuracy.

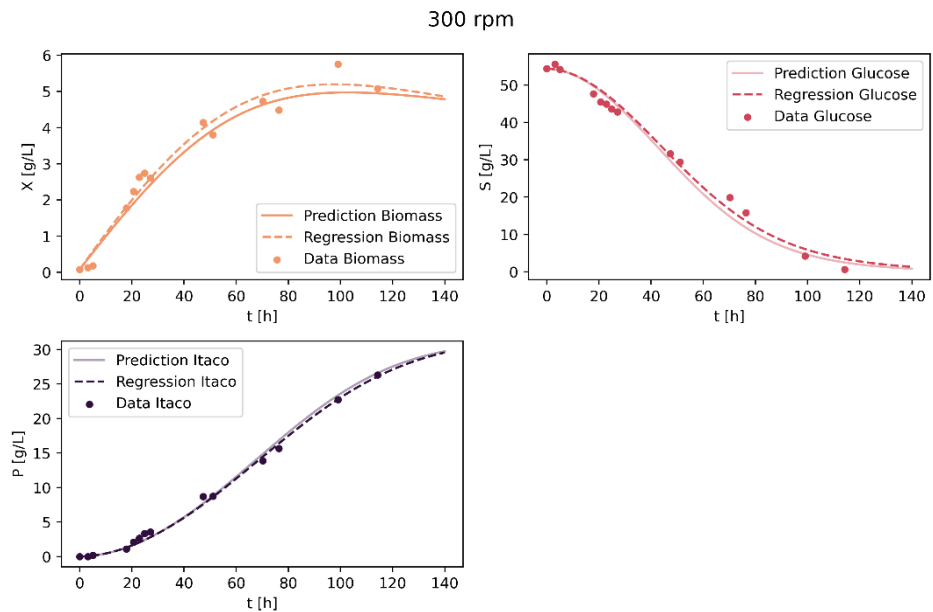

*Figure 6. Comparison of CFM result (solid line) with conventional regression (dashed line). Solutions correspond to the operating conditions with a mixer speed of 300 rpm, an aeration rate of 1 vvm in a 0.5 L reactor.*

*Table 5. Parameter values obtained for experimental data (114 and 229 rpm, 1 vvm, 42 L reactor).*

|  | Parameter | $k_d$ [$h^{-1}$] | $\mu_m$ [$h^{-1}$] | $K_S$ [$g/L$] | $Y_{XS}^{-1}$ [$L/g$] | $k_p$ [$g/(L \cdot h)$] | $\tau_s$ [$h$] | $K_{PS}$ [$g/L$] | $K_X$ [$g/L$] |
|---|---|---|---|---|---|---|---|---|---|
| 114 rpm | Regression | 0.0061 | 0.1738 | 35.22 | 2.581 | 0.1812 | 29.68 | 50.99 | 30.39 |
|  | DDL | 0.0052 | 0.323 | 83.58 | 3.352 | 0.1294 | 43.95 | 60.32 | 52.28 |
|  | CFM | 0.0053 | 0.2945 | 87.14 | 2.819 | 0.1618 | 41.02 | 62.59 | 54.48 |
| 229 rpm | Regression | 0.0057 | 0.3685 | 19.81 | 1.093 | 0.0874 | 28.58 | 61.89 | 52.79 |
|  | DDL | 0.0051 | 0.4814 | 41.73 | 1.585 | 0.092 | 26.71 | 60.38 | 62.2 |
|  | CFM | 0.0055 | 0.5332 | 38.22 | 1.105 | 0.1045 | 24.83 | 63.34 | 90.0 |

*The parameters were obtained using conventional regression, direct deep learning approach and conditional flow matching.*

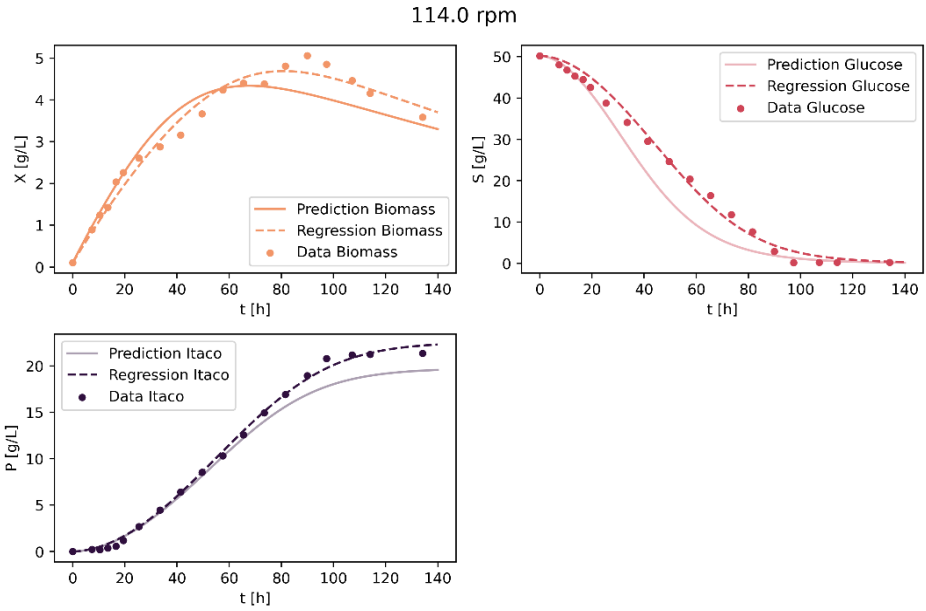

*Figure 7. Comparison of direct deep learning approach result (solid line) with conventional regression (dashed line). Solutions correspond to the operating conditions with a mixer speed of 114 rpm, an aeration rate of 1 vvm in a 42 L reactor.*

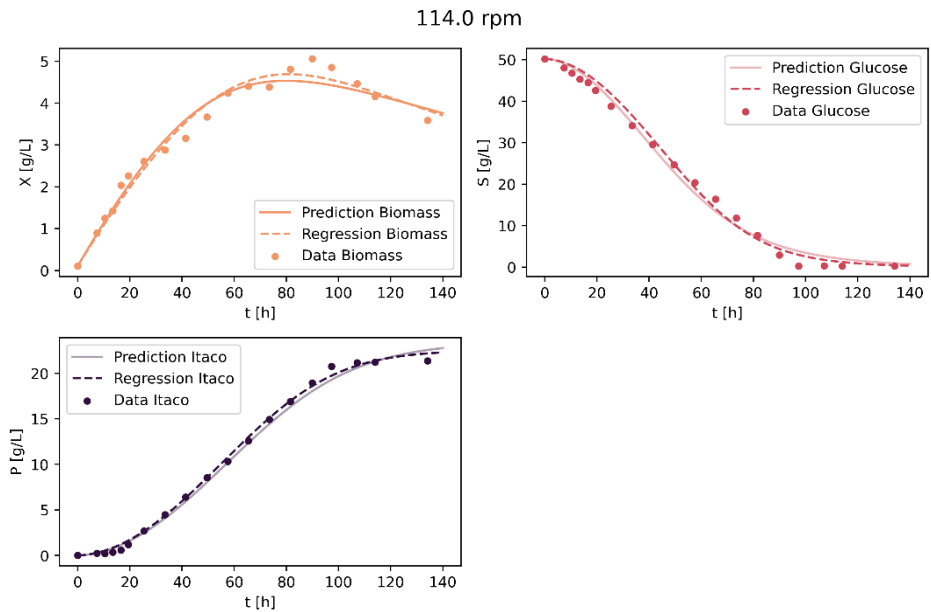

*Figure 8. Comparison of CFM result (solid line) with conventional regression (dashed line). Solutions correspond to the operating conditions with a mixer speed of 114 rpm, an aeration rate of 1 vvm in a 42 L reactor.*

# 5 Conclusion

This work demonstrates the ability of deep learning to estimate model parameters using real experimental data. These learning approaches require only forward simulations as training data, thus avoiding globality and convergence issues often encountered in nonlinear regression. Especially, the results obtained with CFM are comparable to those produced by conventional nonlinear regression, suggesting that CFM is accurate enough for practical bioprocess modeling tasks. In contrast to nonlinear regression, CFM does not suffer from convergence issues or require multistart strategies to avoid poor local minima, which simplifies its application in practice. Compared to DDL, a significant advantage of CFM is its ability to produce good approximations from limited training data while yielding information on probability distributions, thus enabling uncertainty estimates.

At the same time, it should be noted that neural network training requires a certain amount of computational effort. In this study this took approximately seven minutes on a GPU node for a dataset of 500 samples. This time is comparable to the time needed to estimate parameters for several experiments using nonlinear regression. Importantly, for the learning approaches, the training step only needs to be performed once. Afterward, parameter predictions for new datasets can be generated almost instantaneously, enabling rapid analysis in ongoing or future experiments.

Overall, the results indicate that CFM offers a flexible, data-efficient, and robust framework for parameter estimation in dynamic bioprocess models. This framework combines competitive accuracy with the ability to produce fast predictions and quantify parameter uncertainty.

# Acknowledgements

The authors acknowledge Daniil Sherki, Ekaterina Muravleva and Ivan Oseledets for introducing the idea of using conditional flow matching for parameter estimation, cf. (Sherki et al. 2025), on which this study builds.

# Declarations

**Funding.** The experimental measurements were prepared within the project "GreenProScale—Process integration and scale-up of a biorefinery for green waste, taking into account the robustness of the system" from the innovation area "BioBall," which is funded by the German Federal Ministry of Research, Technology and Space (BMFTR, grant numbers 031B1497A and 031B1497B).

**Availability of data and materials.** Data is available upon request from the corresponding author. Parts of the data have been previously published in (Volkmar et al. 2025)

# References

Bates DM, Watts DG (1988) Nonlinear regression analysis and its applications. Wiley series in probability and mathematical statistics. Wiley, New York

Becker J, Tehrani HH, Ernst P, Blank LM, Wierckx N (2020) An Optimized Ustilago maydis for Itaconic Acid Production at Maximal Theoretical Yield. J Fungi (Basel) 7. https://doi.org/10.3390/jof7010020

Biegler LT, Zavala VM (2009) Large-scale nonlinear programming using IPOPT: An integrating framework for enterprise-wide dynamic optimization. Computers & Chemical Engineering 33:575–582. https://doi.org/10.1016/j.compchemeng.2008.08.006

Biegler LT (2010) Nonlinear programming: Concepts, algorithms, and applications to chemical processes. MOS-SIAM series on optimization. Society for Industrial and Applied Mathematics; Mathematical Programming Society, Philadelphia

Carstensen F, Klement T, Büchs J, Melin T, Wessling M (2013) Continuous production and recovery of itaconic acid in a membrane bioreactor. Bioresour Technol 137:179–187. https://doi.org/10.1016/j.biortech.2013.03.044

Geiser E, Wiebach V, Wierckx N, Blank LM (2014) Prospecting the biodiversity of the fungal family Ustilaginaceae for the production of value-added chemicals. Fungal Biol Biotechnol 1:2. https://doi.org/10.1186/s40694-014-0002-y

Geiser E, Przybilla SK, Friedrich A, Buckel W, Wierckx N, Blank LM, Bölker M (2016) Ustilago maydis produces itaconic acid via the unusual intermediate trans-aconitate. Microb Biotechnol 9:116–126. https://doi.org/10.1111/1751-7915.12329

Glass DS, Jin X, Riedel-Kruse IH (2021) Nonlinear delay differential equations and their application to modeling biological network motifs. Nat Commun 12:1788. https://doi.org/10.1038/s41467-021-21700-8

Helm T, Stausberg T, Previati M, Ernst P, Klein B, Busche T, Kalinowski J, Wibberg D, Wiechert W, Claerhout L, Wierckx N, Noack S (2024) Itaconate Production from Crude Substrates with U. maydis: Scale-up of an Industrially Relevant Bioprocess. Microb Cell Fact 23:29. https://doi.org/10.1186/s12934-024-02295-3

Hornik K, Stinchcombe M, White H (1989) Multilayer feedforward networks are universal approximators. Neural Networks 2:359–366. https://doi.org/10.1016/0893-6080(89)90020-8

Hosseinpour Tehrani H, Becker J, Bator I, Saur K, Meyer S, Rodrigues Lóia AC, Blank LM, Wierckx N (2019) Integrated strain- and process design enable production of 220 g L-1 itaconic acid with Ustilago maydis. Biotechnol Biofuels 12:263. https://doi.org/10.1186/s13068-019-1605-6

Klement T, Milker S, Jäger G, Grande PM, Domínguez de María P, Büchs J (2012) Biomass pretreatment affects Ustilago maydis in producing itaconic acid. Microb Cell Fact 11:43. https://doi.org/10.1186/1475-2859-11-43

Liebal UW, Ullmann L, Lieven C, Kohl P, Wibberg D, Zambanini T, Blank LM (2022) Ustilago maydis Metabolic Characterization and Growth Quantification with a Genome-Scale Metabolic Model. J Fungi (Basel) 8. https://doi.org/10.3390/jof8050524

Lipman Y, Chen RTQ, Ben-Hamu H, Nickel M, Le M (2023) Flow matching for generative modeling

Monod J (1949) The growth of bacterial cultures. Annu. Rev. Microbiol. 3:371–394. https://doi.org/10.1146/annurev.mi.03.100149.002103

Peebles W, Xie S (2023) Scalable diffusion models with transformers. In: Proceedings of the IEEE/CVF international conference on computer vision, pp 4195–4205

Rackauckas C, Nie Q (2017) DifferentialEquations.jl – A Performant and Feature-Rich Ecosystem for Solving Differential Equations in Julia. The Journal of Open Research Software 5. https://doi.org/10.5334/jors.151

Raponi A, Marchisio D (2024) Deep learning for kinetics parameters identification: A novel approach for multi-variate optimization. Chemical Engineering Journal 489:151149. https://doi.org/10.1016/j.cej.2024.151149

Rensonnet G, Adam L, Macq B (2021) Solving inverse problems with deep neural networks driven by sparse signal decomposition in a physics-based dictionary. https://arxiv.org/pdf/2107.10657

Rihan FA, Tunc C, Saker SH, Lakshmanan S, Rakkiyappan R (2018) Applications of Delay Differential Equations in Biological Systems. Complexity 2018. https://doi.org/10.1155/2018/4584389

Sherki D, Oseledets I, Muravleva E (2025) Combining Flow Matching and Transformers for Efficient Solution of Bayesian Inverse Problems. In: AI4X 2025 International Conference

Vaswani A, Shazeer N, Parmar N, Uszkoreit J, Jones L, Gomez AN, Kaiser L, Polosukhin I (2017) Attention Is All You Need. https://doi.org/10.48550/arXiv.1706.03762

Villadsen J, Nielsen J, Lidén G (2011) Bioreaction Engineering Principles. Springer US, Boston, MA. https://doi.org/10.1007/978-1-4419-9688-6

Volkmar M, Laudensack W, Bartzack F, Erdmann N, Schönrock S, Fuderer E, Holtmann D, Blank LM, Ulber R (2025) Low Oxygen Availability Increases Itaconate Production by Ustilago maydis. Biotechnol Bioeng 122:3007–3017. https://doi.org/10.1002/bit.70035

Voll A, Klement T, Gerhards G, Büchs J, Marquardt W (2012) Metabolic modelling of itaconic acid fermentation with Ustilago maydis. Chemical Engineering Transactions 27:367–372

Wulkow N, Telgmann R, Hungenberg K-D, Schütte C, Wulkow M (2021) Deterministic and Stochastic Parameter Estimation for Polymer Reaction Kinetics I: Theory and Simple Examples. Macro Theory & Simulations 30. https://doi.org/10.1002/mats.202100017

Ziegler AL, Ullmann L, Boßmann M, Stein KL, Liebal UW, Mitsos A, Blank LM (2024) Itaconic acid production by co-feeding of Ustilago maydis: A combined approach of experimental data, design of experiments, and metabolic modeling. Biotechnol Bioeng 121:1846–1858. https://doi.org/10.1002/bit.28693